\documentclass[lettersize,journal]{IEEEtran}
\usepackage{amsmath,amsfonts}
\usepackage{array}
\usepackage[caption=false,font=normalsize,labelfont=sf,textfont=sf]{subfig}
\usepackage{textcomp}
\usepackage{stfloats}
\usepackage{url}
\usepackage{verbatim}
\usepackage{graphicx}
\usepackage{cite}
\usepackage[linesnumbered,ruled,vlined]{algorithm2e}
\SetKwInput{KwInput}{Input}
\SetKwInput{KwOutput}{Output}
\usepackage{balance}
\usepackage{color, xcolor}
\usepackage{multicol}
\begin{document}

\title{Enhancing Federated Learning Convergence with Dynamic Data Queue and Data Entropy-driven Participant Selection}

\author{Charuka Herath,~\IEEEmembership{Student Member,~IEEE,}
Xiaolan Liu,~\IEEEmembership{Member, ~IEEE,}
and Sangarapillai Lambotharan,~\IEEEmembership{Senior Member, ~IEEE,}
Yogachandran Rahulamathavan

\thanks{ 
 C. Herath, X. Liu, S. Lambotharan, and Y. Rahulamathavan are with the Institute for Digital Technologies, Loughborough University London, London, U.K. (e-mails: \{c.herath, xiaolan.liu, s.lambotharan, y.rahulamathavan\}@lboro.ac.uk).
}

% \thanks{This paper was produced by the IEEE Publication Technology Group. They are in Piscataway, NJ.}
% <-this % stops a space
% \thanks{Manuscript received June 4, 2024.}
}

% \markboth{Journal of \LaTeX\ Class Files,~Vol.~14, No.~8, August~2021}



\maketitle

\begin{abstract}
Federated Learning (FL) is a decentralized approach for collaborative model training on edge devices. This distributed method of model training offers advantages in privacy, security, regulatory compliance, and cost-efficiency. Our emphasis in this research lies in addressing statistical complexity in FL, especially when the data stored locally across devices is not identically and independently distributed (non-IID). We have observed an accuracy reduction of up to approximately 10\% to 30\%, particularly in skewed scenarios where each edge device trains with only 1 class of data. This reduction is attributed to weight divergence, quantified using the Euclidean distance between device-level class distributions and the population distribution, resulting in a bias term (\(\delta_k\)). As a solution, we present a method to improve convergence in FL by creating a global subset of data on the server and dynamically distributing it across devices using a Dynamic Data queue-driven Federated Learning (DDFL). Next, we leverage Data Entropy metrics to observe the process during each training round and enable reasonable device selection for aggregation. Furthermore, we provide a convergence analysis of our proposed DDFL to justify their viability in practical FL scenarios, aiming for better device selection, a non-sub-optimal global model, and faster convergence. We observe that our approach results in a substantial accuracy boost of approximately 5\% for the MNIST dataset, around 18\% for CIFAR-10, and 20\% for CIFAR-100 with a 10\% global subset of data, outperforming the state-of-the-art (SOTA) aggregation algorithms.
\end{abstract}

\begin{IEEEkeywords}
Data-entropy, Fairness FL, Federated Learning, non-IID
\end{IEEEkeywords}

\section{Introduction}
\label{Section: Introduction}
\IEEEPARstart{T}{he} remarkable expansion of cloud-based AI solutions has been closely linked to the rapid growth of the Artificial Intelligence (AI) market significantly in the Internet of Things (IoT) related industries and topics. A primary driver of this technological revolution is the increasing prevalence of personal smart devices. These intelligent devices are now an integral part of people's lives, equipped with numerous sensors that provide access to vast amounts of valuable training data crucial for developing machine learning (ML) models \cite{FLOpt2015, ComEff2017}. Considering the issues of data availability, and privacy sensitivity, Federated Learning (FL) has become increasingly popular as it allows organizations to leverage the distributed data available on personal smart devices without the need to centralize it \cite{FLMulti2017, FLhealth2019, FLforMbKey2018}. Moreover, the rapid proliferation of the Internet of Things (IoT) has ushered in an era of interconnected devices generating vast volumes of data across various application domains. In this context, our proposed Dynamic Data queue-driven Federated Learning (DDFL) approach emerges as a promising solution to optimize ML models in IoT scenarios.

FL involves two key entities: the device, responsible for owning the training data, and the server, the aggregator for the primary model (global model) by utilizing each device's gradients. The distinguishing aspect of FL lies in its ability to achieve superior global model performance while keeping all training data securely stored within the devices' devices. FL architecture operates in training rounds. During each round, the server initializes the parameters of the global model and distributes it to each device. devices then train their local models using their respective data. Once training is complete, the devices upload their local models to the server, where an aggregation algorithm \cite{Junchuan2022} is applied to combine and refine the models into an enhanced global model. This collaborative and privacy-conscious approach to ML is at the forefront of research and is making significant strides in various application domains such as personal credit scoring, environmental monitoring, health and gesture monitoring, traffic and closing management in autonomous vehicles, etc.

Yet FL has drawbacks when dealing with real-world data distribution, device selection for secure aggregation, privacy and security. The distribution of data among devices can be classified into two main categories: independent and identically distributed data (IID) and not identically and independently distributed (non-IID). In practical scenarios, private data use for FL is unbalanced and biased towards unique model creations and will result in less model accuracy, and poor convergence. Works by \cite{niidconvergence2020, FLnIIDDataSilos2022} state that the presence of non-IID data distribution introduces a range of complex challenges within the context of FL. In non-IID scenarios, data distributions across devices are disparate and may exhibit covariate shift, concept drift, and imbalanced learning. This diversity often leads to biased models and skewed predictions, undermining fairness, communication overhead and difficulties in aggregating models due to varying optima further delaying model convergence.

\begin{table}[!t]
\caption{Key Features and challenges in existing work concerning the proposed FL Framework for Dynamic non-IID data.\label{Table: Key feature comparison}}
\centering
\scriptsize
\begin{tabular}{|p{2.3cm}|l|l|l|l|l|l|l|}
\hline
\textbf{Feature} & \textbf{\cite{FlNiid2018}}	& \textbf{\cite{FedMD2019}} & \textbf{\cite{icc_conf_client_selection}} & \textbf{\cite{warmup2022}} & \textbf{\cite{Reza2016}} &  \textbf{\cite{defKt2022}} & \textbf{Ours}\\
\hline
High-accuracy model design with non-IID data & \checkmark  & \checkmark &            & \checkmark &          & \checkmark &  \checkmark\\
\hline
Faster model Convergence                     & \checkmark  & \checkmark & \checkmark & \checkmark & \checkmark & & \checkmark\\
\hline
Dynamic Data distribution                    &             &            &            &            &          & & \checkmark\\
\hline
Fairness device selection               &             &            & \checkmark &            &          & & \checkmark\\
\hline
Data entropy-based aggregation               &      	   &     		&			 &            &          & & \checkmark\\
\hline
\end{tabular}
\end{table}

In the context of IID data distribution, the devices possess data independently sampled from the same underlying distribution. This means that the data across all devices are relatively similar in terms of feature distributions, class proportions, and overall characteristics. The IID assumption simplifies the aggregation process during model updates, as the global model can be efficiently learned by averaging the individual device models \cite{FLComEff2016}. In contrast in non-IID setting devices have distinct and heterogeneous data, not conforming to the IID assumption. In the presence of non-IID data, the aggregation of device updates becomes more challenging, as models trained on different data distributions need to be reconciled which results in biased models. In our study, we refer to this bias made by each device to the global model as the bias term $\delta$. Moreover, \cite{FLConApps2019} discussed the effect of non-IID data on the convergence and performance of FL algorithms and proposed methods to address the heterogeneity of device data. Furthermore, they introduced a communication-efficient FL approach that considers non-IID data distribution to improve convergence in heterogeneous environments.
% Moreover,  \cite{FedAvg2017} proposed the Federated Averaging algorithm (FedAvg), which leverages both non-IID and IID data to perform averaging of local models, leading to robust and scalable global model learning.

\begin{figure}[t]
  \centering
  \includegraphics[width=3.2in]{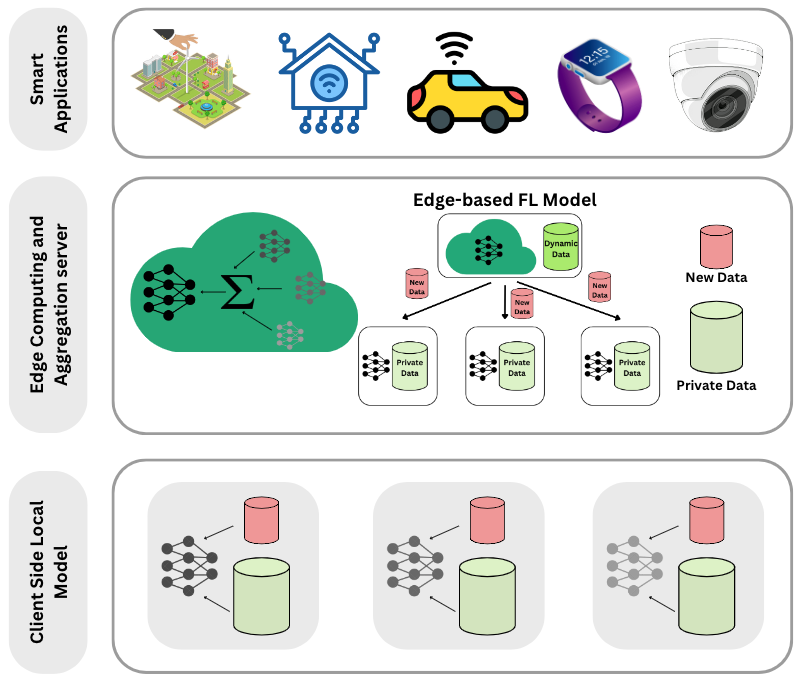}\\
  \caption{Architectural overview of proposed FL for IoT network}\label{Figure: FL_IoT}
\end{figure}

An article by \cite{OnConvFL2018} states that challenges encompass data distribution, thereby delaying the fast convergence of a global model. Additionally, \cite{FLMulti2017} states that the dissimilarity in data distributions necessitates heightened communication between the central server and devices, potentially resulting in communication bottlenecks arising from disparate update frequencies. Furthermore, \cite{FlNiid2018} states that non-uniform distribution can also engender biased model updates, as varying data patterns and class distributions per device influence the learning process. The preservation of privacy and security within this decentralized framework becomes an intricate concern, given the potential for aggregation to accidentally expose sensitive information \cite{FLEdgeInt2023}. Furthermore, the absence of centralized supervision hampers the determination of optimal global model updates. A work by \cite{FLEdge2022} states that accommodating model heterogeneity stemming from non-IID data requires tailored aggregation strategies to ensure effective knowledge fusion while maintaining privacy. \cite{OpenProbFL2019} states that addressing these challenges demands the implementation of personalized aggregation mechanisms, dynamic communication strategies, and specialized techniques like federated meta-learning to foster convergence speed, communication efficiency, and bias mitigation, all while upholding stringent privacy standards. Moreover, the device selection in FL involves addressing data fairness, bias, and the diverse capabilities of heterogeneous devices. It's crucial to select devices in a way that represents a fair cross-section of the data, avoiding biases that could skew the model, especially in non-IID data scenarios. Additionally, considering the varying capabilities of devices ensures an equitable contribution to the global model, optimizing the learning process without overburdening any particular device. Balancing these aspects in device selection is key to achieving effective and fair FL \cite{FLOptNiid2020}. \textbf{Table. \ref{Table: Key feature comparison}} states the challenges in existing work in brief and a detailed comparison of the challenges can be found in \textbf{Section \ref{Background and Related Work}}.

Building upon the principles of FL, DDFL addresses the challenges posed by non-IID data, a common issue in IoT ecosystems. As illustrated in Fig. \ref{Figure: FL_IoT}, this work finds relevance in diverse IoT applications such as smart cities, healthcare, and industrial IoT. For instance, in smart cities, where heterogeneous data streams from sensors and devices abound, DDFL ensures improved model accuracy and convergence. In healthcare IoT, patient data from disparate sources can be effectively utilized with fairness and robustness through DDFL. Moreover, in industrial IoT, predictive maintenance models benefit from DDFL's capacity to adapt to dynamic data distributions across machinery. The issues of communication efficiency, privacy preservation, and non-IID data distribution in IoT align with the strengths of DDFL \cite{infoEntopy2013, PPTraficIoT2020, ComEffIoT2020}.

Understanding the implications of both IID and non-IID data distribution is critical for designing efficient and effective FL algorithms. Addressing the challenges posed by non-IID data, such as devising adaptive aggregation methods and personalized model updates, is essential for optimizing the performance of federated models across diverse and distributed data sources. In this study, we propose a robust method of high-accuracy model design by introducing a novel approach to handling non-IID data which prioritizes data entropy-based aggregation and fairness in device selection for non-IID settings. Moreover, \textbf{Table. \ref{Table: Key feature comparison}} elaborates the key features of our proposed approach and compares them with the state-of-the-art (SOTA) approaches which will be discussed in \textbf{Section. \ref{Background and Related Work}}.

%%%%%%%%%%%%%%%%%%%%%%%%%%%%%%%%%%%%%%%%%%

\subsection{Contributions and Motivations}

Our method is designed to tackle several critical challenges commonly encountered in the training process, including slow convergence and low model accuracy. Moreover, we focus on the efficient selection of devices for updating the global model, a key aspect that leads to a more equitable and fair model training process in FL by reducing the biases between the global model and local models. To summarize, our work makes the following significant contributions to the field.

\begin{enumerate}
    \item \textbf{Faster convergence and better accuracy:} We propose a novel data distribution mechanism for non-IID data where each device in each round has additional training data. In this approach, the $\gamma$ portion of data will be saved in the server and distributed dynamically among devices in each training round. It will enhance the model convergence speed, average model aggregation time and accuracy. Here we replicate the real-world FL scenario where the local model training data is dynamic. Additionally, we observe a reduction in the bias term $\delta$, quantified by the Euclidean distance between device-level class distributions and the population distribution, resulting in improved model convergence.
    \item \textbf{Rigourous experiment:} We comprehensively evaluate the issue of non-IID data which leads to poor accuracy, sub-optimal models and low convergence using model weight divergence. We further compare the weight divergence concerning the DDFL approach and evaluate the DDFL approach using data entropy.
    \item \textbf{Entrophy-based user section:} We propose a model aggregation mechanism by effectively selecting $\lambda$ devices based on the data entropy of each local device. This approach will ensure the fairness of device selection during aggregation. Our analytical results show that in each training round the data entropy improved significantly. As a result, the effect on data in each round and devices rich in data will play a vital role in model convergence and accuracy in FL.
\end{enumerate}

\subsection{Paper Organisation}

The subsequent sections of this paper are organized as follows. \textbf{Section \ref{Background and Related Work}} summarizes the related work in the same research areas and the challenges in the current work. \textbf{Section \ref{Section: Problem Statement and Formulation}} states the problem scope and provides a mathematical explanation of why the traditional FedAvg fails on non-IID data. Furthermore, shows the accuracy reduction and low convergence rate when having non-IID data, and explains the root cause using weight divergence in each scenario and the distribution of classes in each device. \textbf{Section \ref{Section: Proposed Methodology}} explains the proposed DDFL Framework. \textbf{Section \ref{Section: Experiments and Evaluation}} states the results and discussion on the proposed approach which reduces $\delta$, improves the convergence speed and improves model accuracy by ~10\% and efficient model convergence. \textbf{Section \ref{Section: Conclusion}} serves as the conclusion of the paper, summarizing the key findings, contributions and future work.

%%%%%%%%%%%%%%%%%%%%%%%%%%%%%%%%%%%%%%%%%%

\section{Background and Related Work}
\label{Background and Related Work}

% A study in \cite{Reza2016} proposed the FedSGD method, in which for all devices, the gradient will be sampled in fractions. In the global server, gradients will be scaled proportionally to device samples and are aggregated.

\subsection{The Preliminaries of Federated Learning}

In the context of FL, the goal is to reduce the federated objective function $F(\mathbf{w})$, where $w$ denotes the model parameters. This objective function, $F(\mathbf{w})$, is formulated as the weighted mean of individual local objective functions $F_k(\mathbf{w})$ corresponding to each device $k$ within the FL framework. Mathematically, we have:

\begin{equation}
\underset{w}{\min} ~ F(\mathbf{w}), \text{ where } F(\mathbf{w}) = \sum_{k=1}^{K} p_k F_k(\mathbf{w}),
\end{equation}

\noindent where $K$ signifies the overall number of devices, with $p_k \geq 0$ and the constraint that $\sum_{k=1}^{K} p_k = 1$. The function $F_k(\mathbf{w})$ denotes the local objective function corresponding to the $k^{th}$ device. Typically, this local objective function is defined as the empirical risk based on the data specific to that particular device as in the following formula

\begin{equation}\label{local objective function}
F_k(\mathbf{w}) = \frac{1}{\hat{x}_k} \sum_{j=1}^{\hat{x}_k} f(w, \hat{x}_{k,j}, \hat{y}_{k,j}),
\end{equation}

\noindent In this context, $\hat{x}_k$ stands for the quantity of locally available samples on device $k$, and $(\hat{x}_{k,j}, \hat{y}_{k,j})$ denotes the $j^{th}$ sample along with its associated label. The parameter $p_k$ serves as a user-defined weight, determining the proportional influence of each device's contribution on the overall objective function. Two common settings for $p_k$ are $p_k = \frac{n_k}{\sum_{k=1}^{K} n_k}$ or $p_k = \frac{\hat{x}_k}{n}$, where $\hat{x} = \sum_{k=1}^{K} \hat{x}_k$ is the total number of samples across all devices.

The FedAvg, as outlined by \cite{FedAvg2017} in \textbf{Algorithm \ref{Algorithm: FedSGD}}, represents an enhanced iteration of FedAvg and stands out as a cutting-edge aggregation mechanism for FL. This approach enables devices to train their local models using diverse sets of local data. Subsequently, these models are exchanged with the central server.

\begin{algorithm}[!ht]
\DontPrintSemicolon
\tcc{System Initialisation}
    \KwInput{Locally trained models ${F_1(\mathbf{w})}$, ${F_2(\mathbf{w})}$,..., ${F_k(\mathbf{w})}$ from each devices for the $n^{th}$ epoch}
    \KwOutput{Global model $F(\mathbf{w})^*$}
\tcc{Learning}
\For{$k \leq K $}
    {
         Add each local model together while averaging by partition data used by ${k^{th}}$ device \\
         {
            Average the sum of models \\
            $F(\mathbf{w}) = \sum_{k=1}^{K} p_k F_k(\mathbf{w})$
         }
    }
\label{Algorithm: FedSGD}
\caption{FedAvg}
\end{algorithm}

\subsection{Related Work}

A work by \cite{FlNiid2018} proposed a practical data-sharing strategy which involves a globally shared dataset, denoted as $G_D$, which is centrally stored in the cloud. This dataset $G_D$ comprises a uniform distribution over classes. The process begins with the initialization phase of FedAvg, where a warm-up model trained on $G_D$ and a random $\gamma$ proportion of $G_D$ are allocated to each device. Each device possesses a local model, which is trained not only on the shared data from $G_D$ but also on the private data unique to each device. The local model training leverages a combination of shared data and device-specific data to learn and improve its performance. Once the local models are trained on their respective data, the cloud performs model aggregation using FedAvg. This aggregation process involves combining the local models from all devices to create a global model that represents the collective knowledge from the entire network of devices. There are two key trade-offs to consider in this data-sharing strategy which are the trade-off between the size of the globally shared dataset $G_D$ and the test accuracy. 

This trade-off is quantified by the parameter $\gamma$, which is the percentage of the size of $G_D$ ($||G_D||$) to the total data from all devices ($||L_D||$) expressed as a percentage ($\gamma = \frac{||G||}{||L_D||} \times 100\%$). Adjusting the size of $G_D$ can influence the overall test accuracy of the FL system. Secondly, there is a trade-off between the proportion $\gamma$ and the test accuracy of the globally shared dataset $G_D$ distributed to each device during the initialization stage. The value of $\gamma$ determines the fraction of shared data available to each device at the beginning of the training process. Different values of $\gamma$ can impact the final test accuracy achieved by the FL model. 
% Finding the right balance between these two trade-offs is essential to optimize the performance of the FL system while considering the constraints on data privacy, communication costs, and computation resources. Overall, this data-sharing strategy in FL seeks to strike a balance between data sharing and privacy preservation. However, it will not address the dynamic data distribution issue in FL. 

Additionally, in \cite{warmup2022}, a novel technique was introduced involving utilising a warmup model. This approach entails the initial model being capable of leveraging pre-trained weights and adjusting to the data similarity among clients through the feature extractor. The weights generated by the warmup model serve the purpose of mitigating the effects of weight divergence within the models.

The work by \cite{FedMD2019} introduces the concept of ``model distillation" as a cornerstone of their approach. The framework named FL via Model Distillation (FedMD) is presented to harness model distillation to facilitate proficient learning in scenarios characterized by heterogeneous data distributions. This approach is designed to concurrently uphold the imperatives of privacy and communication efficiency. At the core of model distillation lies the principle of training a more compact student model to emulate the performance of a larger, proficient teacher model. The authors underscore the limitations faced by conventional aggregation techniques such as Federated Averaging when confronted with non-IID data. They emphasize that the straightforward averaging process employed in traditional methods can result in compromised convergence rates and sub-optimal outcomes. Furthermore, in \cite{icc_conf_client_selection}, they presented Client Selection for FL (FedCS), which focuses on managing devices efficiently by considering their resource conditions. This protocol effectively addresses a device selection problem, accounting for resource constraints. By doing so, the server can aggregate a maximum number of device updates, thereby accelerating the performance enhancement of ML models. Moreover, \cite{fairness_guarantee} proposed a method that selects devices with a fairness guarantee in training. Furthermore, the experimental results show that a fairer strategy could promise efficiency in training and model convergence and higher final accuracy. However, all these approaches perform below average in non-IID data distribution.

A work in \cite{du2022gradient} proposed a dynamic device scheduling mechanism, which can select qualified edge devices to transmit their local models with a proper power control policy to participate in the model training at the server for FL via over-the-air computation. Moreover, work in \cite{minJoingLer2019} proposed an adaptive device selection mechanism for FL. Devices are selected based on their communication channel conditions and computational resources. This approach ensures that devices with better connectivity and computational capabilities contribute more to the model training process, leading to faster convergence and improved performance. Moreover, \cite{FLChallenges2020} introduced a priority-based device scheduling mechanism for FL. Devices are prioritized based on the importance of their updates to the global model. The measure of importance is determined by considering factors such as the relevance of the local data and the impact of the device's update on the model's performance. This approach accelerates model convergence and enhances overall performance by prioritising devices with more significant updates. Furthermore, the work in \cite{FedML2020} proposed a dynamic client selection approach using reinforcement learning. The selection of clients for participation in FL is treated as a sequential decision-making process. Reinforcement learning techniques dynamically adapt client selection based on channel conditions, update importance, and historical performance. This adaptive approach improves the efficiency and effectiveness of FL in heterogeneous environments. Moreover, in \cite{defKt2022}, the author proposed a novel mutual knowledge transfer algorithm called Def-KT in a decentralized federated learning (FL) setup. Traditional FL involves the participating clients sending their model updates to the central server after each round.

Although each of these approaches is promising, we have found that they are plagued by common issues such as the computational overhead caused by complex system designs, low accuracies when handling complex and high-dimensional datasets like CIFAR-100, and biased device selection. In essence, DDFL is a novel approach that addresses the most pressing challenges in FL: slow convergence, local model accuracy, and effective and fair client selection based on the quality of each device's data, even when dealing with complex and high-dimensional datasets in a non-IID data setting.
%%%%%%%%%%%%%%%%%%%%%%%%%%%%%%%%%%%%%%%%%%
\begin{table}[t]
    \centering
    \scriptsize
    \caption{Table of Notations}
    \centering
    \begin{tabular}{|c|c|c|c|}
    \hline
    $F(\mathbf{w})$ & Objective Function & $F_k(\mathbf{w})$ &  Local Objective Functions\\
    \hline
    $k$ & Local Device & $\hat{x}_k$ & Local sample Qty\\
    \hline
    $K$ & Total Local Devices & $\eta$ & Learning Rate\\
    \hline
    $d_k(\hat{x})$ & non-IID Distribution & $\mathcal{L}$ & Loss Function\\
    \hline
    ${H(\chi)}$ & Data Entropy & $G_D$ & Dynamic Data set \\
    \hline
    $N$ & Global Epochs & $F(\mathbf{w})^*$ & Global Model \\
    \hline
    $n$ & Current Epoch & $F_k^n(\mathbf{w})$ & Local Model \\
    \hline
    $\delta_k$ & Bias term & $\gamma$ & Dynamic Data Potion\\
    \hline
    $\lambda$ & Aggregation devices proportion & $b$& batch size\\
    \hline
    $\zeta_k$ & reliability index & $\sigma_n$ & SD of test accuracy\\
    \hline
    $\mu_n$ & mean of test accuracy & $p_k$ & normalization of weights \\
    \hline
    \end{tabular}
    \label{Table: Table of Notations}
\end{table}

%%%%%%%%%%%%%%%%%%%%%%%%%%%%%%%%%%%%%%%%%%

\section{Problem Statement and Formulation}
\label{Section: Problem Statement and Formulation}

In this section, we demonstrate that the reasons for challenges in FL: slow convergence, and local model accuracy, are due to non-IID data leading to sub-optimal models. Furthermore, extensive experiment results will be discussed in \textbf{Section \ref{Section: Experiments and Evaluation}}.

\subsection{non-IID Causing on Sub-optimal Global Model}

Suppose a client possesses significantly more data than others or has a significantly high amount of data from a single class following a non-IID distribution. In that case, it can bias the global model towards its skewed distribution, diverting performance on other devices’ data. This is a critical drawback of the weight aggregation strategy, which may lead to a sub-optimal model inference.

Now, let's analyze the impact of the weights \(p_k\) on the model updates which were previously discussed in \textbf{Section \ref{Background and Related Work}} which is deriving from Equation (\ref{local objective function}). The SGD update for FL is typically performed as follows:

\begin{equation}
w^{(n)}_k = w^{(n-1)}_{k} - \eta \nabla F_k(w^{(n-1)}_{k})
\end{equation}

Assuming a small learning rate \(\eta\), the update term can be approximated as:

\begin{equation}
\Delta w^{(n)}_k \approx - \eta \nabla F_k(w^{(n-1)}_{k})
\end{equation}

Now, consider the impact of \(p_k\) on the update:

\begin{equation}
\Delta w^{(n)}_k \approx - \eta p_k \nabla F(w^{(n-1)}_{k})
\end{equation}

where \(\nabla F(w^{(n-1)}_{k})\) is the gradient of the global objective function concerning the parameters at device \(k\).

If \(p_k\) is chosen based on the local data size, the update is influenced by \(p_k\). Devices with larger data sizes (\(x_k\)) will have smaller \(p_k\) values, leading to larger updates. This can result in biased model updates, favouring devices with larger data sizes. If \(p_k\) is chosen as \(p_k = \frac{x_k}{x}\) (normalized by the total data size across all devices), the bias in model updates is mitigated to some extent. However, bias can still occur if the local data distributions are highly skewed.

In both cases, if the \(p_k\) values do not accurately reflect the true distribution of data characteristics across devices, bias in model updates can lead to a global model that is skewed towards the data characteristics of devices with higher weights. Therefore, careful consideration of the choice and normalization of weights (\(p_k\)) is crucial to minimize bias in model updates and ensure fair contributions from all participating devices in FL.

Furthermore, Let $F(k)$ represent the globally optimal model that we would achieve if we could aggregate all the data in a centralized manner. In FL with non-IID data, each device $k$ computes a local model $F_k(\mathbf{w})$ based on its respective non-IID data distribution $d_k(\hat{x})$.

In the local model update the local model of device \(k\) is obtained by minimizing its local loss function:
\begin{equation}
    {F(\mathbf{w}) = \arg\min_{F_k(\mathbf{w})} \mathcal{L}_k(F_k(\mathbf{w}))}
\end{equation}
where \(\mathcal{L}_k(\mathbf{w})\) is the local loss function based on the non-IID data distribution \(d_k(x)\). Then the global model is updated by aggregating the local models:
\begin{equation}
    {F(\mathbf{w}) = \sum_{i=k}^{K} \frac{\hat{x}_k}{\hat{x}} F_k(\mathbf{w})}
\end{equation}
where \(K\) is the total number of devices, \(\hat{x}_k\) is the size of device \(k\)'s data, and \(\hat{x}\) is the total size of the aggregated data. Define a bias term \(\delta_k\) representing the bias introduced by device \(k\):
\begin{equation}
    {\delta_k = F_k(\mathbf{w}) - F(\mathbf{w})}
\end{equation}

The updated global model considers bias from all devices:
\begin{equation}
    {F(\mathbf{w})^* = F(\mathbf{w}) + \sum_{k=1}^{K} \frac{\hat{x}_k}{\hat{x}} \delta_k}
\end{equation}

Due to the non-IID nature of the data, $F(\mathbf{w})^*$ might not be the same as the globally optimal model $F(\mathbf{w})$ that we would obtain with IID data, which leads to a sub-optimal global model $F(\mathbf{w})$.

\subsection{Effect of non-IID on Slower Convergence Speed}

Due to non-IID data, the updates proposed by each device $k$ during FL have different convergence rates. This is due to the varying data distributions and quality across devices, leading to diverse learning rates $\eta$. In a typical FL update, the global model $F(\mathbf{w})$ at round $n$ is updated as follows:

\begin{equation}
  F(\mathbf{w})^* = F(\mathbf{w}) - \eta \nabla L(F_(\mathbf{w}), \hat{x}_k), \quad \text{where } \hat{x}_k \sim d_k(\hat{x})
\end{equation}

devices with more representative or easier-to-learn data (higher quality) may have larger learning rates, converging faster. Conversely, devices with challenging or less representative data may have smaller learning rates, converging more slowly. The consequence of diverse learning rates is that model updates across devices have varying convergence speeds. This leads to a slower overall convergence of the global model since we need to aggregate these diverse updates over multiple communication rounds.

Various strategies have been proposed to address non-IID data distribution challenges in FL. The approach by \cite{FlNiid2018} involves a globally shared dataset ($G_D$) for initialization, with trade-offs between the size of $G_D$ and test accuracy, and between $\gamma$ and the accuracy of $G_D$ during initialization. Additionally, \cite{warmup2022} introduces a warmup model technique to address weight divergence effects. The work by \cite{FedMD2019} uses ``model distillation," while \cite{icc_conf_client_selection} presents FedCS, managing devices efficiently based on resource conditions. \cite{fairness_guarantee} proposes a fairness guarantee method. Thus, \cite{defKt2022} is a promising approach, the design architecture leads to computational and communications overhead. Despite their contributions, these approaches exhibit limitations in addressing issues like sub-optimal global models, slow convergence, and biased device selection in FL with non-IID data. To overcome these challenges, our proposed DDFL and novel aggregation mechanism based on devices' data entropy aim to enhance convergence, accuracy, and fairness in dynamic real-world scenarios.

%%%%%%%%%%%%%%%%%%%%%%%%%%%%%%%%%%%%%%%%%%
%%%%%%%%%%%%%%%%%%%%%%%%%%%%%%%%%%%%%%%%%%

\section{Proposed DDFL Framework}
\label{Section: Proposed Methodology}

In this section, we explain our proposed method to overcome the challenge of non-IID data distribution in dynamic real-world scenarios by introducing the DDFL and the novel aggregation mechanism based on the data entropy of devices. A work by \cite{infoEntopy2013} proposed the concept of information entropy which can be seen as a measure of the average amount of information needed to describe an event or message. In information theory, higher entropy implies higher uncertainty and a greater need for information to represent the data. Conversely, lower entropy indicates a more predictable or structured dataset, requiring less information for its representation.
In mathematical representation, Entropy can be defined as a discrete random variable ${H(\chi)}$, which takes items of certain classes $\hat{x}$ in a dataset and distributes them such that  ${p: \hat{x} \rightarrow [0, 1]}$. Here, in the following equation, the negative sign ensures that the overall entropy is a positive value or zero, implying that lower entropy corresponds to more ordered or certain datasets.

\begin{equation}\label{equation_entropy}
    { H(\chi) = - \sum_{i} p(\hat{x}_k) \cdot \log_2(p(\hat{x}_k))}
\end{equation}

\subsection{Dynamic Data Distribution - DDFL}

In our study, in MINIST and CIFAR-10 experiments $i$ = 10 as there are 10 classes in both datasets and $i$ = 100 for the CIFAR-100 dataset.
In the IID setting, each device is randomly assigned a uniform distribution over the 10 classes. In the non-IID setting, the data is organized by class and partitioned, creating an extreme scenario of single-class non-IID, wherein each device exclusively receives data from a unique class. The data distribution is explained in \textbf{Fig. \ref{Figure: Data Entropy Distribution}} using data entropy in the \textbf{Section \ref{Section: Experiments and Evaluation}}.

For both the non-IID and IID settings, we uphold a queue housing $\gamma$ = 0.1 training samples. These samples are randomly selected from this queue in each training iteration to cater to every device's learning process. Importantly, during this phase, the class assignments are randomized, introducing an additional layer of complexity to the data composition. Furthermore, we set $\gamma$ = 0.2 as an alternative setting setting. A deep analysis of the experimental results and the justifications for selecting effective hyper-parameters are stated in \textbf{Table. \ref{Table: Accuracy Comparison}}.

\begin{algorithm}[t]
\caption{Dynamic Data Distribution and Aggregation (DDFL)}
\label{Algorithm: DDD}
\KwData{Dynamic dataset $G_D$, Number of devices $K$, Epochs $N$}
\KwResult{Improved and Robust global model $F(\mathbf{w})^*$}
\For{$n = 1$ to $N$}{
    \tcc{Initialize global model $F(\mathbf{w})^*$ with baseline structure and distribute it among devices}
    \For{$k = 1$ to $K$}{
        Receive fresh randomly data segment $D_{k_i}$ from $G_D$\;
        Train local model $F_k(\mathbf{w})$ using data $D_k$ and $D_{k_i}$\;
        Calculate local update $F_k^n(\mathbf{w})$\;
        \tcc{Calculate the Data Entropy using Equation \ref{equation_entropy}}
        ${H(\chi) = - \sum_{i} p(\hat{x}_k) \cdot \log_2(p(\hat{x}_k))}$\;
    }
    Receive Local Updates ($F_k^n(\mathbf{w})$...$F_K^n(\mathbf{w})$) and Data Entropies (${H(\chi_k)}$...${H(\chi_K)}$)\;
    \tcc{Aggregate device updates using Algorithm \ref{Algorithm: Aggregation}}
    \textit{Aggregation}(${[{F_k^n(\mathbf{w})}...{F_K^n(\mathbf{w})}], [{H(\chi_k)}...{H(\chi_K)}]}$)\;
}
\end{algorithm}

\begin{figure}[t]
  \centering
  \includegraphics[width=3.2in]{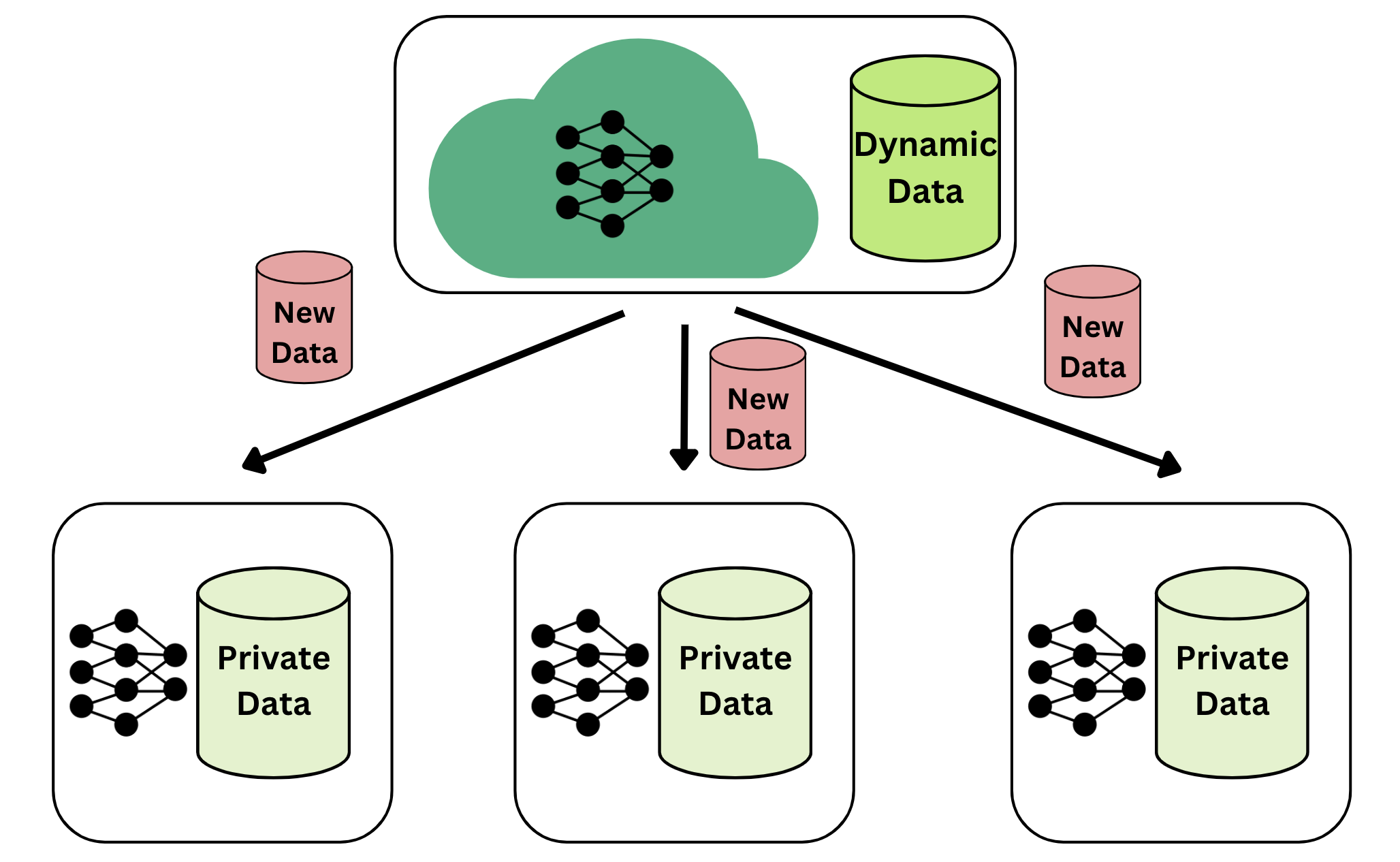}\\
  \caption{Proposed Federated Learning Framework (DDFL)}\label{Figure: Data Distribution Method}
\end{figure}

\begin{figure}[ht]
    \centering
    \subfloat[Non-Dynamic]{\includegraphics[width=1.5in]{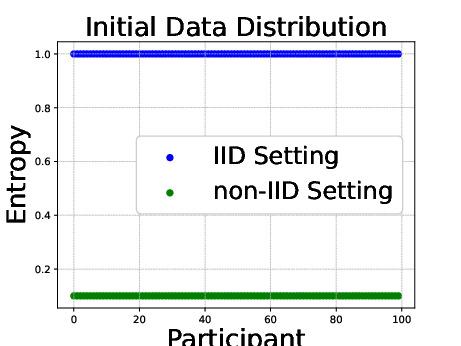}
    }
    \hfil
    \subfloat[Dynamic]{\includegraphics[width=1.9in]{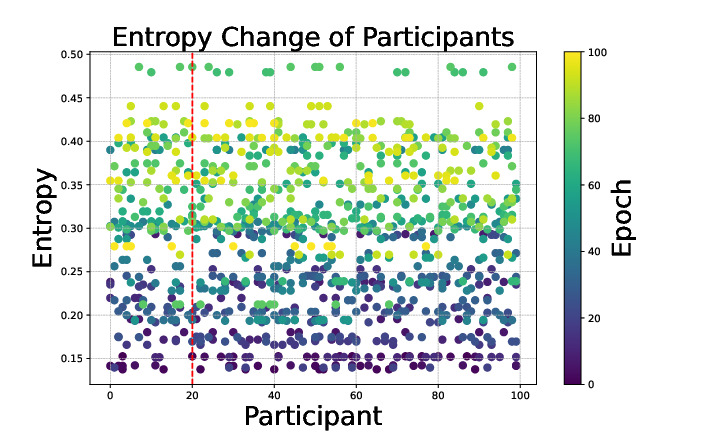}}
    \caption{Fig. (a) shows the initial Data Entropy in IID and non-IID data Distributions. Fig. (b) shows the data Entropy distribution for devices in each training round in the proposed DDFL. The colour bar is introduced to the respective epoch.}
    \label{Figure: Data Entropy Distribution}
\end{figure}

As illustrated in \textbf{Algorithm \ref{Algorithm: DDD}} we introduce a novel data-sharing approach within the context of FL, as depicted in \textbf{Fig. \ref{Figure: Data Distribution Method}}. We establish a centralized dataset denoted as $G_D$, which is stored in a global data queue and comprises a uniform distribution spanning all available classes in datasets. For this experiment, we stored $10\%$ data samples as $G_D$ and the rest of the $90\%$ were distributed among the devices. This dataset resides in the server. At the onset of the initialization phase, the baseline model structure is shared with the participating entities. Starting from the second iteration onwards, each device $k$ is allocated a segment of fresh data $D_{k_i}$ extracted from the global dataset $G_D$. 

% \begin{figure}[t]
%   \centering
%   \includegraphics[width=3.2in]{figures/FL_IoT.png}\\
%   \caption{Architectural overview of proposed FL for IoT network}\label{Figure: FL_IoT}
% \end{figure}

Moreover, \textbf\textbf{Fig. 3b}, illustrates the data entropy variation of each device in each training round in the proposed method. for example in the $20^{th}$ device's data entropy will be changed from \~0.1 to \~0.48 during 100 epochs. It is an increment of 80\% total increment for the data entropy.

Subsequently, devices train their respective local models using the \textit{FedAvg}. Throughout this training, devices' models are fine-tuned based on their assigned data subsets. In each training round, devices calculate the local update $F_k^n(\mathbf{w})$ and data entropy $H(\chi_k)$ calculated using Equation (\ref{equation_entropy}) and send it to the server.

\begin{equation}
    F(\mathbf{w})^* = \frac{\sum\limits_{k=1}^K H(\chi_k)F_k^n(\mathbf{w})}{\sum\limits_{k=1}^K H(\chi_k)}
\end{equation}

For instance, we consider the $n^{th}$ synchronization round. Here, we have $k$ number of local updates and compare their respective, $H(\chi_k)$ and select $\lambda$ amount of devices for aggregation. Furthermore, we conclude that this potion of devices has the highest quality data from all $k$ devices in the $n^{th}$ round. Moreover, we ran out simulations for an extensive amount of rounds and decided the best $\gamma$ value is 0.9. In \textbf{Section \ref{Section: Experiments and Evaluation}} we illustrate our test results and justify selecting this hyper-parameter. Then each device's updates will be scaled based on their data entropies using Equation (\ref{equation_entropy}).

Finally, the global model will be averaged using the scaled device updates. This mechanism ensures fairness in FL when it comes to selecting devices that have low skewness in data.

\subsection{Entropy-based Aggregation}

The concluding step in this iterative process involves the server executing an aggregation mechanism, specifically designed according to the aggregation algorithm outlined in \textbf{Algorithm \ref{Algorithm: Aggregation}}. In this context, we focus on selecting the top $\lambda$ fraction of devices with the highest entropy values during each training round and utilize their computed gradients for the aggregation process. For our experiments and evaluations, we set the baseline $\lambda$ value to 0.9 and compared it with an alternative setting where $\lambda$ was set to 0.8, as detailed in \textbf{Section \ref{Section: Experiments and Evaluation}}. By consolidating the updates provided by each device, this aggregated knowledge is subsequently employed to enhance the overall performance of the global model.

\begin{algorithm}[t]
\caption{Aggregation Function}
\label{Algorithm: Aggregation}
\SetKwProg{Fn}{Function}{:}{}
\Fn{Aggregation(${[{F_K^n(\mathbf{w})}], [{H(\chi_K)}]}$)}{
    Select $\lambda$ of the devices $K$ with heights ${H(\chi)}$\;
    \For{$k = 1$ to $K$}{
        Average the ${k^{th}}$ device's model\;
        ${F_k^n(\mathbf{w})} = H(\chi_k){F_k^n(\mathbf{w})}$\;
    }
    Global model aggregation\;
    $F(\mathbf{w})^* = \frac{\sum\limits_{k=1}^K H(\chi_k)F_k^n(\mathbf{w})}{\sum\limits_{k=1}^K H(\chi_k)}$
}
\end{algorithm}

The next section will illustrate our simulations and findings from the proposed framework.

\subsection{Convergence Analysis}

In this section, we perform fundamental convergence analysis for our proposed DDFL algorithm, the server selects the top $\lambda$ fraction of devices which have higher data entropy for aggregation. We analyze the convergence performance of the proposed DDFL algorithm under non-IID data \cite{niidconvergence2020} with the dynamic device selection scheme. We first present the preliminaries and assumptions, and then the convergence result is obtained.
\subsubsection{Preliminaries}
The optimal solution of the global loss function \( L(\mathbf{w}) \) is defined in the \textbf{Equation (6)}.
% \begin{equation}
%     {F(\mathbf{w}) = \arg\min_{F_k(\mathbf{w})} \mathcal{L}_k(F_k(\mathbf{w}))}
% \end{equation}
So the minimum loss is \( L^* = L(F(\mathbf{w})) \). Similarly, the minimum loss of $k$, \( L_k \) is denoted by \( L^*_k = L_k(F(\mathbf{w})) \). Then the local-global objective gap is defined as
\begin{equation}
\label{degree_of_nIID}
\Phi = L^* - \sum_{k=1}^{K} p_k L^*_k,
\end{equation}
where \( \Phi \) is nonzero, quantifying the degree of non-IID data; its magnitude reflects the heterogeneity of the data distribution. Larger $\Phi$ implies higher data heterogeneity over the K. If the data is IID and the number of samples is large then $\Phi \rightarrow 0$ [7].
\subsubsection{Assumptions}
We make the following assumptions on the functions $F_1, \cdots, F_N$. Assumptions 1 and 2 are standard; typical examples are the $\ell_2$-norm regularized linear regression, logistic regression, and softmax classifier.

\textbf{Assumption 1.} $F_1, \cdots, F_N$ are all L-smooth: for all $\mathbf{v}$ and $\mathbf{w}, F_k(\mathbf{v}) \leq F_k(\mathbf{w})+(\mathbf{v}-$ $\mathbf{w})^T \nabla F_k(\mathbf{w})+\frac{L}{2}\|\mathbf{v}-\mathbf{w}\|_2^2$.

\textbf{Assumption 2.} $F_1, \cdots, F_N$ are all $\mu$-strongly convex: for all $\mathbf{v}$ and $\mathbf{w}, F_k(\mathbf{v}) \geq F_k(\mathbf{w})+(\mathbf{v}-$ $\mathbf{w})^T \nabla F_k(\mathbf{w})+\frac{\mu}{2}\|\mathbf{v}-\mathbf{w}\|_2^2$.
\noindent Assumptions 3 and 4 have been made by the works \cite{comeff_2013, stich_2018}.

\textbf{Assumption 3.} Let $\xi_n^k$ be sampled from the $k$-th device's local data uniformly at random. The variance of stochastic gradients in each device is bounded: $\mathbb{E}\left\|\nabla F_k\left(\mathbf{w_n^k}, \xi_n^k\right)- \nabla F_k\left(\mathbf{w_n^k}\right)\right\|^2 \leq \sigma_k^2$ for $k=1, \cdots, N$.

\textbf{Assumption 4.} Suppose the samples on all the devices are homogeneous due to the dynamic environment assume that $N \rightarrow \infty$. In that case, i.e., they are sampled in an IID fashion, then as $H(\chi_k) \rightarrow 1$, $p_k \rightarrow \frac{1}{K}$,  it follows 
$\mid\sigma_{k,n}\mid > \mid\sigma_{k,n+1}\mid$.

\textbf{Assumption 5.} Assume that that assumption 4 holds,  it follows $\mid\sigma_{i,n} - \sigma_{j,n}\mid  > \mid\sigma_{i,n+1} - \sigma_{j,n+1}\mid, (i, j \subset K)$.

\textbf{Assumption 6.} The expected squared norm of stochastic gradients is uniformly bounded, i.e., $\mathbb{E}\left\|\nabla F_k \left (\mathbf{w_n^k}, \xi_n^k\right)\right\|^2 \leq G^2$ for all $k=1, \cdots, K$ and $n=0, \cdots, N-1$

Here we analyze the case in which the $\lambda K$ device participates in the aggregation step. Let the FedAvg algorithm terminate after $N$ iterations and return $w_N$ as the solution. We always require $N$ to be evenly divisible by $E$ so that FedAvg can output $w_N$ as expected.

\textbf{Theorem 1.} Let Assumptions 1 to 6 hold and $L, \mu, \sigma_k, G, E$ (Steps of the local Update) be defined therein. Choose $\kappa = \frac{L}{\mu}$, $\varrho = \max\{8\kappa, E\}$ and the learning rate $\eta_n = \frac{2}{\mu(\varrho + n)}$. Then FedAvg with spatial device participation satisfies \cite{Li_2019},

\[
\mathbb{E}\left[F\left(\mathbf{w}_N\right)\right]-F^* \leq \\ 
\frac{ \kappa}{\varrho+N - 1}\left(\frac{2B}{\mu}+ \frac{\mu\varrho}{2} \mathbb{E} \left\|\mathbf{w}_0-\mathbf{w}^*\right\|^2\right) 
\]
Where,
\begin{equation}
\label{Equation: B}
    B = \sum_{k=1}^N p_k^2 \sigma_k^2 + 6L\Phi + 8(E - 1)^2 G^2 + \frac{4}{K}E^2G^2
\end{equation}

\textbf{Assumption 7} Suppose the samples on all the devices are homogeneous and let Assumption 4 hold. In that case, i.e., they are sampled in an IID fashion, then as $H(\chi_k)$, it follows $p_k \rightarrow \frac{1}{K}$,  it follows that $\Phi \rightarrow 0$ for every \textbf{w} as all the local functions converge to the same expected risk function in the large sample limit.

\textbf{Theorem 2.} DDFL requires $p_k = \frac{H(\chi_k)}{\sum_{k=1}^{K} H(\chi_k)}$; $p_1 + \dots + p_K = 1$ which does not violate the unbalanced nature of FL. Let $F_{k}(w) \simeq p_k N \bar{F}_k(w)$ be a scaled local objective $F_k$. Let Assumption 7 hold, then the global objective becomes a simple average of all scaled local objectives:

Then the global objective becomes a simple average of all scaled local objectives:
\begin{equation}
\label{Equation: Objective Function}
F(\mathbf{w})=\sum_{k=1}^{\lambda K} p_k F_k(\mathbf{w}) \simeq  \frac{1}{K} \sum_{k=1}^{K} \bar{F}_{k}(w)
\end{equation}

The derived proofs of key assumption are shown in \cite{Li_2019}. The
above demonstration proves that DDFL aggregation is guaranteed to converge in FL. The convergence speed is $O\left(\frac{1}{N}\right)$. B in Equation (\ref{Equation: B}) will shrink based on Assumptions 3,4 and 7 and it will satisfy the objective in Equation (\ref{Equation: Objective Function}).

\section{Experiments and Evaluation}
\label{Section: Experiments and Evaluation}

\subsection{Experimental Setup}
The experiments were conducted on a high-performance computing setup, utilizing an NVIDIA RTX 6000 GPU with 48GB of VRAM, coupled with an Intel Core i9-10980 processor. This hardware configuration ensured efficient training and evaluation of our models, providing reliable and indicative results.

In the experimental setup,  we utilized three distinct datasets, MNIST, CIFAR-10, and CIFAR-100 to evaluate the performance of our models. MNIST comprises 60,000 training images and 10,000 test samples, each depicting handwritten grayscale digits ranging from 0 to 9 in a 28x28 format. CIFAR-10 consists of 60,000 32x32 colour images distributed across 10 classes. In contrast, CIFAR-100 consists of 60,000 32x32 colour images, just like CIFAR-10, but is divided into 100 distinct classes. Each class contains 600 images, with 500 images for training and 100 images for testing. For MNIST, a Convolutional Neural Network (CNN) was employed, featuring two convolution layers with 32 and 64 kernels of size 3 × 3, followed by a max-pooling layer and two fully connected layers with 9216 and 128 neurons. Meanwhile, CIFAR-10 and CIFAR-100 involved a more intricate model architecture, specifically the ResNet-18, chosen to accommodate the complexity of the dataset with its 10 classes and 100 class cases. This experimental design allowed us to comprehensively assess the models' capabilities across varied image classification challenges.

\subsection{Dynamic Data Distribution}

In our proposed DDFL approach, we computed data entropy during the initial epoch for both non-IID and IID settings in this section, using the CIFAR-10, CIFAR-100 and MNIST datasets. Moreover, we calculated the dynamic data entropy for each device in each synchronization round. Equation (\ref{equation_entropy}) guided our computation of individual data entropies of each setting.

As anticipated, the entropy visualization in \textbf{Fig. 3a} validates the concept that the IID approach yields a data entropy of \textbf{1} for each device. This outcome results from the uniform distribution of classes among participating entities. Moreover, \textbf{Fig. 3b} confirms the expected data entropy of \textbf{0.1} for all devices in the non-IID setting. This is because each device has data from a single class, highlighting the absence of uniform data distribution.

In contrast, when employing dynamic data distribution, device data entropies evolve. \textbf\textbf{Fig. 3b} demonstrates that the data entropy of each device increases with the increasing of epochs, leading to improved data distribution uniformity. In training epoch 1, the data entropy of each device is 0.1, akin to the initial non-IID distribution. Then, it's evident that the data entropy significantly increases from the $1^{st}$ training round to the $100^{th}$ training round. Nonetheless, class imbalances persist within each local dataset as the data is distributed randomly.

In the DDFL approach, device selection is determined by data entropy. Notably, the data entropy for each device increases with each epoch. For example, as illustrated in \textbf\textbf{Fig. 3b}, during the $20^{th}$ epoch, devices contributing to the aggregation are those with the highest data entropy. This mechanism serves to eliminate bias and unfairness in device selection within FL. This statement will be justified in the next section using the simulation results. 

\subsection{The detailed performance of the proposed framework}
During simulations, we compared the proposed approach in the non-IID setting with the FedAvg \cite{Reza2016} in non-IID and IID data settings and the Warmup model \cite{warmup2022} in the non-IID setting by considering them as unique SOTA methods. 

\begin{figure*}[ht]
    \centering
    \includegraphics[width=7in]{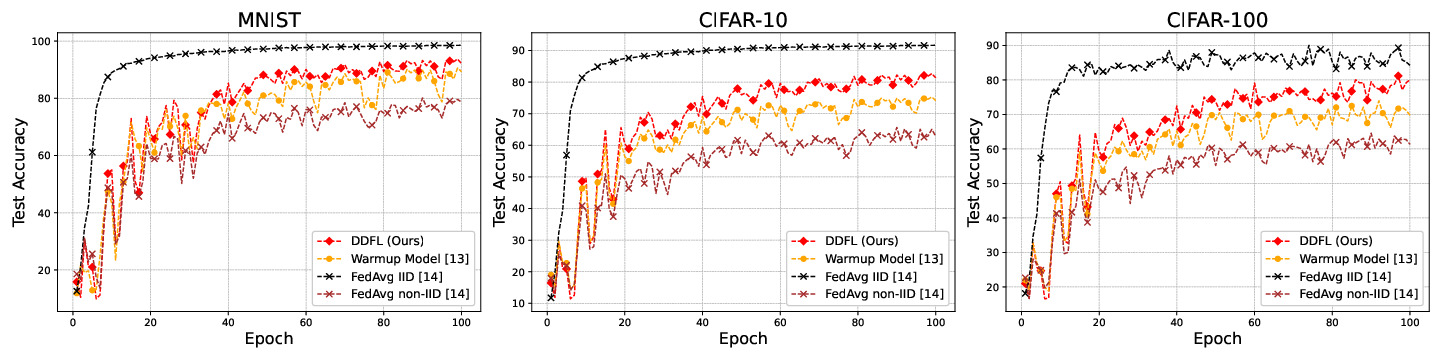}
    \caption{Figures show the Test accuracy over the epoch of \textit{FedAvg} with IID and non-IID setting, DDFL, and the warmup model with the non-IID setting for MNIST, CIFAR-10 and CIFAR-100 datasets.}
    \label{Figure: DDFL final}
\end{figure*}

% \begin{figure*}[ht]
%     \centering
%     \subfloat[CIFAR-10]{\includegraphics[width=3.5in]{figures/Dynamic_MNIST.jpeg}%
%     \label{Figure: Dynamic MNIST_results}}
%     \hfil
%     \subfloat[MNIST]{\includegraphics[width=3.5in]{figures/Dynamic_CIFAR.jpeg}%
%     \label{Figure: Dynamic CIFAR_results}}
%     \caption{Fig. (a) and Fig. (b) show the Test accuracy over the epoch of \textit{FedSGD} with IID and non-IID setting, DDFL, and the warmup model with the non-IID setting for MNIST and CIFAR-10 datasets.}
%     \label{Figure: DDFL final}
% \end{figure*}

In the non-IID setting, the data is organized based on their respective classes and then divided into an extreme scenario known as 1-class non-IID. In this extreme scenario, each device exclusively receives data from just one class. We investigate three approaches by adjusting the value of $\gamma$ to 0.1 and 0.2, and $\lambda$ to 0.8, 0.9, and an extreme 0.5.

As depicted in \textbf{Fig. \ref{Figure: DDFL final}}, under the IID data sets, all three datasets exhibit high accuracies, and the model achieves convergence between 20-40 epochs. Conversely, in a dynamic non-IID setting, employing initial FedAvg and the warm-up model approaches reveals a notable decline in accuracy and sub-optimal model convergence. The accuracies for MNIST datasets range between 78\% - 85\%, for the CIFAR-10 dataset, they range between 63\% - 74\%, and for the CIFAR-100 dataset, they range between 59\% - 70\%  under these approaches.
However, a distinct improvement is evident with the implementation of the proposed DDFL approach. The accuracy increases by a minimum of 4\% - 10\% for MNIST, 6\% - 18\% for CIFAR-10, and 18\% - 20\% for CIFAR-100 datasets. Moreover, the model demonstrates accelerated convergence, particularly notable between epochs 40 - 50.

\begin{table*}[ht]
    \centering
    \caption{Experimental results under different hyper-parameters $b=100$, $e$, $\gamma$ and $\lambda$ for MNIST, CIFAR-10, and CIFAR-100 datasets (DD - Data Distribution, AAT - Average Aggregation time in s) Accuracies in \% in each setting are displayed}.
    \label{Table: Accuracy Comparison}
    \newcolumntype{C}{>{\centering\arraybackslash}X}
    \scriptsize
    \begin{tabular}{c|cccccc|cccccc} \hline
        \textbf{Approach} & \multicolumn{6}{c}{ \textbf{Approach 01 $e$ = 5 $\gamma$ = 0.1  $\lambda$ = 0.8}} \vline & \multicolumn{6}{c}{ \textbf{Approach 02 $e$ = 1 $\gamma$ = 0.1 $\lambda$ = 0.9}} \\
        \textbf{DD} & \textbf{MNIST} & \textbf{AAT} & \textbf{CIFAR-10} & \textbf{AAT} & \textbf{CIFAR-100} & \textbf{AAT} &  \textbf{MNIST} & \textbf{AAT} & \textbf{CIFAR-10} & \textbf{AAT} & \textbf{CIFAR-100} & \textbf{AAT}\\ \hline
        IID \cite{FlNiid2018} & 98.51 & 1.65 & 93.61 & 2.69 & 77.98 & 3.85 & 98.56 & 1.66 & 93.74 & 2.71 & 78.13 & 3.89 \\ 
        non-IID (baseline) \cite{FlNiid2018} & 89.30 & 1.95 & 65.19 & 3.49 & 57.64 & 4.86 & 89.53 & 1.96 & 66.93 & 3.57 & 56.59 & 4.83\\
        Def-KT \cite{defKt2022} & 91.32 & 1.92 & 75.74 & 3.54 & 62.43 & 5.69 & 91.45 & 2.41 & 75.12 & 3.66 & 63.74 & 5.34\\ 
        warm-up  \cite{warmup2022} & 90.12 & 1.88 & 74.78 & 3.29 & 67.03 & 4.59 & 90.76 & 1.89 & 74.97 & 3.30 & 67.56 & 4.61\\ 
        DDFL (Ours) & 92.17 & 1.86 & 82.34  & 3.16 & 69.59 & 4.53 & \textbf{92.85} & 1.86 & \textbf{83.08} & 3.16 & 70.00 & 4.58\\ \hline
        Accuracy Boost & \multicolumn{2}{c}{ 4.49\%}  & \multicolumn{2}{c}{ 17.78\%} & \multicolumn{2}{c}{ 19.76\%} \vline & \multicolumn{2}{c}{ 4.34\%} & \multicolumn{2}{c}{19.49\%} & \multicolumn{2}{c}{ 22.12\%}\\ 
        AAT Boost & \multicolumn{2}{c}{ 4.61\%} & \multicolumn{2}{c}{ 9.45\%} & \multicolumn{2}{c}{ 7.28\%} \vline & \multicolumn{2}{c}{ 5.10\%} & \multicolumn{2}{c}{ 11.48\% } & \multicolumn{2}{c}{ 5.45\%} \\
        \multicolumn{11}{c}{} \\ \hline
        
        \textbf{Approach} & \multicolumn{6}{c}{ \textbf{Approach 03 $e$ = 5 $\gamma$ = 0.2  $\lambda$ = 0.8}} \vline & \multicolumn{6}{c}{ \textbf{Approach 04 $e$ = 5 $\gamma$ = 0.2 $\lambda$ = 0.9}} \\ 
        \textbf{DD} & \textbf{MNIST} & \textbf{AAT} & \textbf{CIFAR-10} & \textbf{AAT} & \textbf{CIFAR-100} & \textbf{AAT} &  \textbf{MNIST} & \textbf{AAT} & \textbf{CIFAR-10} & \textbf{AAT} & \textbf{CIFAR-100} & \textbf{AAT}\\ \hline
        IID \cite{FlNiid2018} & 98.23 & 1.59 & 92.94 & 2.35 & 78.02 & 3.80 & 98.38 & 1.61 & 93.22 & 2.39 & 78.91 & 3.86\\ 
        non-IID (baseline) \cite{FlNiid2018} & 88.14 & 1.84  & 63.45 & 3.23 & 57.92 & 4.72 & 89.08 & 1.92  & 64.99 & 3.27 & 57.01 & 4.81\\
        Def-KT \cite{defKt2022} & 90.21 & 1.89 & 74.91 & 3.31 & 62.82 & 5.71 & 90.68 & 2.31 & 74.68 & 3.60 & 63.57 & 5.18\\
        warm-up \cite{warmup2022} & 89.8 & 1.69 & 75.67 & 3.09 & 67.19 & 4.37 & 90.07 & 1.74  & 75.92 & 3.13 & 69.68 & 4.45 \\
        DDFL (Ours) & 91.87 & 1.63 & 81.82 & 2.92 & 69.71 & 4.48 & 92.03 & 1.98 & 81.98 & 3.12 & \textbf{71.45} & 4.62\\ \hline
        Accuracy Boost & \multicolumn{2}{c}{ 5.73\%}  & \multicolumn{2}{c}{ 18.53\%} & \multicolumn{2}{c}{ 21.09\%} \vline & \multicolumn{2}{c}{2.93\%} & \multicolumn{2}{c}{16.97\%} & \multicolumn{2}{c}{ 19.11\%}\\
        AAT Boost & \multicolumn{2}{c}{ 5.79\%} & \multicolumn{2}{c}{ 9.59\%} & \multicolumn{2}{c}{ 7.28\%}\vline & \multicolumn{2}{c}{4.89\%} & \multicolumn{2}{c}{5.23\%} & \multicolumn{2}{c}{5.51\%}\\
        \multicolumn{11}{c}{} \\ \hline
        
        \textbf{Approach} & \multicolumn{6}{c}{ \textbf{Approach 05 $e$ = 5 $\gamma$ = 0.5  $\lambda$ = 0.8}} \vline & \multicolumn{6}{c}{ \textbf{Approach 06 $e$ = 1 $\gamma$ = 0.5 $\lambda$ = 0.9}} \\
        \textbf{DD} & \textbf{MNIST} & \textbf{AAT} & \textbf{CIFAR-10} & \textbf{AAT} & \textbf{CIFAR-100} & \textbf{AAT} &  \textbf{MNIST} & \textbf{AAT} & \textbf{CIFAR-10} & \textbf{AAT} & \textbf{CIFAR-100} & \textbf{AAT}\\ \hline
        IID \cite{FlNiid2018} & 96.34 & 1.24 & 90.06 & 2.13 & 63.64 & 2.88 & 96.67 & 1.24 & 90.16 & 2.17 & 64.89 & 2.89 \\
        non-IID (baseline) \cite{FlNiid2018} & 81.25 & 1.64 & 60.27 & 2.94 & 48.78 & 3.53 & 81.89 & 1.68 & 61.04 & 2.98 & 49.53 &3.57  \\ 
        Def-KT \cite{defKt2022} & 83.48 & 1.63 & 63.47 & 3.01 & 52.34 & 4.31 & 84.56 & 1.72 & 66.21 & 3.45 & 59.29 & 4.01\\
        warm-up  \cite{warmup2022}& 82.08 & 1.59 & 62.75 & 2.81 & 55.69 & 3.45 & 83.12 & 1.61 & 62.98 & 2.83 & 56.02 & 3.47 \\
        DDFL (Ours) & 88.76 & \textbf{1.57} & 72.02 & \textbf{2.79} & 57.87 & \textbf{3.44} & 89.08 & 1.58 & 73.89 & 2.81 & 58.79 & 3.42 \\ \hline
        Accuracy Boost & \multicolumn{2}{c}{8.46\%} & \multicolumn{2}{c}{16.31\%} & \multicolumn{2}{c}{ 17.58\%} \vline & \multicolumn{2}{c}{8.07\%} & \multicolumn{2}{c}{17.37\%} & \multicolumn{2}{c}{ 18.42\%}\\
        AAT Boost & \multicolumn{2}{c}{4.26\%} & \multicolumn{2}{c}{5.10\%} & \multicolumn{2}{c}{ 2.61\%} \vline & \multicolumn{2}{c}{5.95\%} & \multicolumn{2}{c}{0.57\%} & \multicolumn{2}{c}{4.38 \%}\\
    \end{tabular}
\end{table*}

The outcomes of our simulations are briefly presented in \textbf{Table. \ref{Table: Accuracy Comparison}}. To delve deeper into our findings, we conducted a comparative analysis of accuracy gain in 6 contrasting scenarios while comparing the DDFL with SOTA FedAvg \cite{FlNiid2018}, Def-KT \cite{defKt2022}, and warmup model \cite{warmup2022}. In the conventional non-IID setting, across the provided combinations, we observed a notable increment in accuracy. Specifically, in the baseline non-IID approach applied to the MNIST dataset, there was an approximate 5\% increase in accuracy, in the case of the CIFAR-10 dataset, the boost was around 18\%, and for CIFAR-100 it was around 20\%. Notably, when $\gamma$ was elevated to 0.2 under $\lambda$ values of 0.8 and 0.9, a marginal decrease in accuracy, approximately 1\%, was noted compared to the prior approach. Moreover in the extreme $\lambda = 0.5$ scenario, in both settings, we were able to observe a severe accuracy decline across all three datasets. However, still, the DDFL approach was able to maintain the accuracy boosts between 17.50\% and 18.50\%. Notably, we observed that the average aggregation time in  $\lambda = 0.5$ is at its lowest as the potion of the training data across devices is lower than the previous approaches. Furthermore, as depicted in the \textbf{Fig. \ref{Figure: DDFL final}}, we can observe the slowness of convergence in this approach. As indicated in the table, the optimal configuration for our experiments materialized when $\gamma$ equalled 0.1, and $\lambda$ was set to 0.9. Furthermore, we observed a significant improvement in average aggregation time (5.10\% in MNIST, 11.48\% in CIFAR-10, and 5.45\% in CIFAR-100) compared to SOTA, in all 6 scenarios. This is a robust matrices to conclude that, DDFL has comparatively low computational overhead due to the simpler design of the aggregation mechanism, dynamic data distribution, and increasing data quality for each device over the epoch. In conclusion, the proposed DDFL overcame all the SOTA approaches while maintaining low aggregation time, higher convergence rate, and higher accuracy. 

\subsection{Determination of Optimal Hyper-parameters}

The choice of Approach 02 $(e = 1, \gamma = 0.1, \lambda = 0.9 )$ as the optimal hyper-parameter combination is based on its strong performance across the used datasets, balancing communication efficiency, model convergence, and adaptability to non-IID data \cite{FlNiid2018}. Furthermore, setting $e = 1$ ensures that models are frequently updated across devices, which is crucial in FL scenarios to reduce the impact of data heterogeneity and mitigate weight divergence. The low $\gamma = 0.1$ value indicates that only 10\% of the data is globally shared, preserving privacy while still enabling some level of data sharing to improve model generalization. A high device selection ratio $\lambda = 0.9$ ensures that most devices participate in each round, fostering a more inclusive and representative global model. For MNIST, with this set of parameters, DDFL achieves an accuracy of 92.85\% with an AAT of 1.86s. On CIFAR-10, the same set of parameters attains 83.08\% accuracy with a 3.16s AAT, and on CIFAR-100, it achieves 70.00\% accuracy with a 4.58ss AAT.

However, this combination needed an adjustment for CIFAR-100, where there are more classes and a higher risk of model divergence due to the non-IID nature of the data \cite{Sahu_2018}. In such cases, setting $e = 5$ was beneficial to allow more local iterations and better local model accuracy before aggregation. Additionally, setting $\gamma = 0.2$ provides a larger shared data subset, which can help in stabilizing training and improving generalization across the diverse classes in CIFAR-100. Thus, for CIFAR-100, an optimal combination of parameters $(e = 5, \gamma = 0.2, \lambda = 0.9)$ allows for more robust training while maintaining the benefits of frequent updates and high device participation. This approach achieves the highest accuracy of 71.45\% for CIFAR-100 with an AAT of 4.62s. Thus, while Approach 02 is optimal for general use, Approach 04 should be employed for complex datasets like CIFAR-100 to achieve superior accuracy and reasonable aggregation times.

Next, we justify the selection of the batch size $b=100$. To ensure a fair comparison across datasets, we selected a base batch size of 100. This choice was motivated by several factors: firstly, a batch size of 100 strikes a balance between computational efficiency and model stability, allowing for efficient training while still providing meaningful batch statistics \cite{batchsize2022}. Furthermore, a work by \cite{LiuBatchSize2020} states that the selection of the batch size needs to be based on the lowest communication and computational costs when the data distribution is non-IID in wireless communication in the FL environment.

\begin{table*}
\centering
\caption{Comparison of test-accuracy results (non-IID data distributions)}
\label{Table:comparison_accuracy_results}
\small
\begin{tabular}{l|cc|cc|cc}
    \hline
    \textbf{} & \multicolumn{2}{c}{\textbf{MNIST}} \vline & \multicolumn{2}{c}{\textbf{CIFAR-10}} \vline & \multicolumn{2}{c}{\textbf{CIFAR-100}} \\
    \textbf{B} & \textbf{FedAvg} & \textbf{DDFL} & \textbf{FedAvg} & \textbf{DDFL} & \textbf{FedAvg} & \textbf{DDFL} \\
    \hline
    10 & 0.91 $\pm$ 0.0026 & 0.93 $\pm$ 0.0074 & 0.65 $\pm$ 0.0028 & 0.81 $\pm$ 0.0031 & 0.34 $\pm$ 0.0052 & 0.66 $\pm$ 0.0052 \\
    25 & 0.90 $\pm$ 0.0029 & 0.93 $\pm$ 0.0028 & 0.66 $\pm$ 0.0008 & 0.82 $\pm$ 0.0024 & 0.51 $\pm$ 0.0053 & 0.67 $\pm$ 0.0054 \\
    50 & 0.89 $\pm$ 0.0067 & 0.92 $\pm$ 0.0095 & 0.66 $\pm$ 0.0048 & 0.83 $\pm$ 0.0002 & 0.58 $\pm$ 0.0048 & 0.68 $\pm$ 0.0027 \\
    \textbf{100} & 0.89 $\pm$ 0.0053 & \textbf{0.92} $\pm$ 0.0085 & 0.66 $\pm$ 0.0097 & \textbf{0.83} $\pm$ 0.0008 & 0.56 $\pm$ 0.0059 & \textbf{0.70} $\pm$ 0.0041 \\
    200 & 0.85 $\pm$ 0.0083 & 0.91 $\pm$ 0.0058 & 0.67 $\pm$ 0.0052 & 0.84 $\pm$ 0.0050 & 0.59 $\pm$ 0.0070 & 0.70 $\pm$ 0.0059 \\
    \hline
\end{tabular}
\end{table*}

\begin{figure*}
    \centering
    \includegraphics[width=7in]{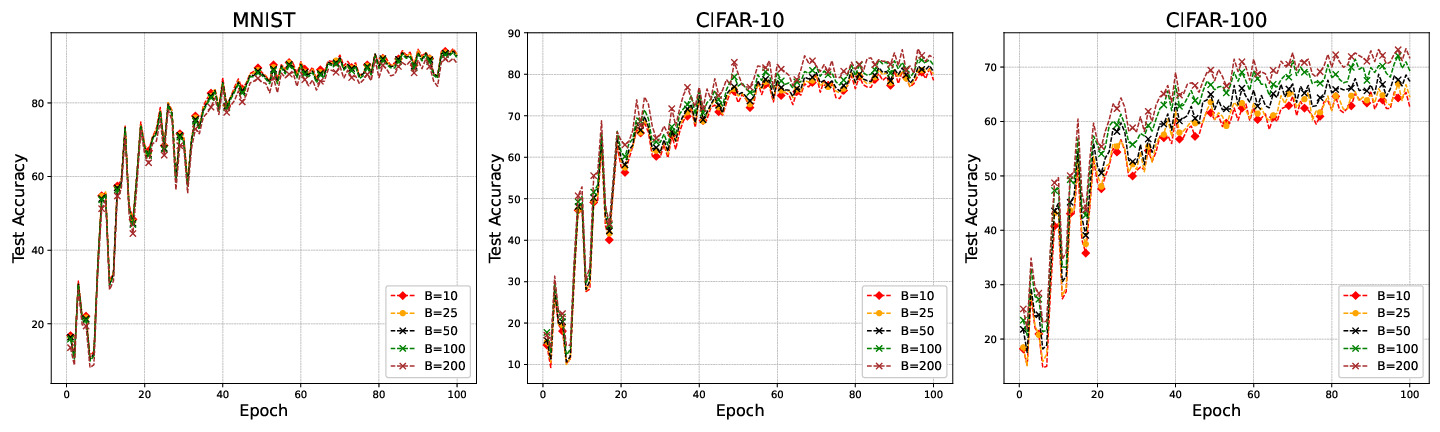}
    \caption{Comparison of accuracies of DDFL approach against MNIST, CIFAR-10 and CIFAR-100 datasets under different batch sizes}
    \label{fig: Comparison accuracy results ac}
\end{figure*}

\begin{figure*}
    \centering
    \includegraphics[width=7in]{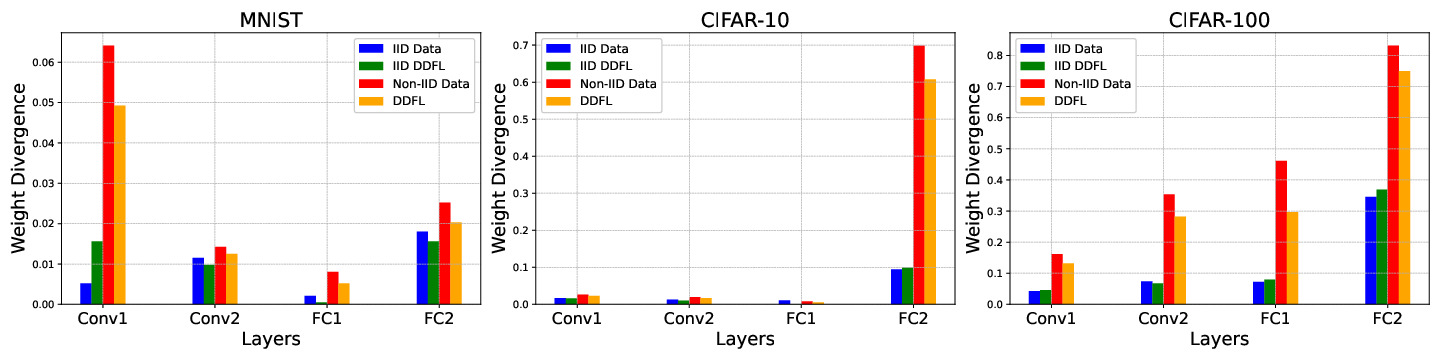}
    \caption{Figures illustrates the Weight divergence of CNN layers for IID, Dynamic IID and Dynamic non-IID for MNIST, CIFAR-10 and CIFAR-100 Datasets}
    \label{Figure: weightDivergence}
\end{figure*}

Additionally, a larger batch size leads to more stable gradient updates and helps mitigate the noise introduced by non-IID data, which is particularly important when dealing with higher-dimensional datasets like CIFAR-100. Despite the increased complexity of the dataset, we observe that our DDFL approach demonstrates robust convergence, outperforming traditional FedAvg across all three datasets. As depicted in the \textbf{Table. \ref{Table:comparison_accuracy_results}}, Specifically, with Approach 02 ($b$ = 100, $e$ = 1, $\gamma$ = 0.1, $\lambda$ = 0.9), (this choice ensures broad participation across devices, preventing overfitting to any single device’s data). These results underscore the effectiveness of DDFL in handling datasets of varying complexities, showcasing its superior performance compared to traditional FedAvg. As a result in this study, we fixed $b$ = 100 across all 6 approaches.

Furthermore, we observed that the average aggregation time is slightly decreased in the proposed approach whereas in non-IID data settings with FedAvg. As displayed in \textbf{Table. \ref{Table: Accuracy Comparison}} the aggregation time dropped at least by \textbf{5\%}, \textbf{10\%}, and \textbf{6\%} when using MNIST, CIFAR-10, CIFAR-100 datasets respectively in both approaches. This suggests that adjusting the value of $\gamma$ can have an impact on the accuracy of the model in extreme non-IID scenarios.

\subsection{Theoretical Grantee using Weight Divergence Comparison}
As stated by \cite{FairFL2019  } Weight divergence in FL refers to the phenomenon where the model parameters or weights of local models trained on different device devices deviate significantly. When device data distributions are non-IID, the updates may not be directly compatible with each other, leading to divergence in the model weights and resulting in slow model convergence, reduced accuracy, and compromised generalization performance. Here we calculate the weight divergence in the last epoch to check the model weight divergence from each local model weight after training.

The Euclidean Distance is a widely used metric to measure the similarity or dissimilarity between two vectors in a multi-dimensional space. In the context of FL, it helps quantify the difference between the weights of the global model; ($F(\mathbf{w})$) and the model of a specific device ($F_k(\mathbf{w})$). In the FL context, comparing the Euclidean Distance across multiple devices allows for an evaluation of how much the models have diverged or converged during the FL process. It serves as a valuable tool for assessing the degree of model consistency across devices and understanding the convergence dynamics of the FL algorithm. It is named weight divergence. The weight divergence is calculated using the Equation (\ref{Equation: weight divergence}). Here, for each parameter (weight) $j$, we calculate the squared difference between the corresponding parameters of the global model $F(\mathbf{w})^*$ and the device's model $F_k^n(\mathbf{w})$, take the sum of square different and finally take the square root.

\begin{equation}
\label{Equation: weight divergence}
    \text{Weight Divergence} = \sqrt{\sum_{j=1}^{N} \left(F(\mathbf{w})^*_j - F_k^n(\mathbf{w})_j\right)^2}
\end{equation}

The Weight Divergence yields a single numerical value that signifies the overall difference between the model parameters of the global model and the device's model. A smaller Weight Divergence implies a lesser difference, indicating that the models are more similar in terms of their weights. Conversely, a larger Weight Divergence suggests a more significant difference between the models.

As depicted in \textbf{Fig. \ref{Figure: weightDivergence}}, weight divergence across every layer rises with the transition from Dynamic IID to non-IID to Dynamic non-IID data. This suggests an inherent connection between weight divergence and data skewness. The accuracy degradation observed in \textbf{Table. \ref{Table: Accuracy Loss for MNIST, CIFAR-10 and CIFAR-100}} can be related to weight divergence, which measures the weight disparity among four distinct training methods with identical weight initialization. As shown in \cite{FlNiid2018} utilized death mover distance for weight divergence analysis, concluding that such divergence arises from the discrepancy between the distribution of data and the skewed distribution of the population of each device. 

\begin{table*}
    \centering
    \label{Table: system_reliability_index}
    \caption{Comparison for model reliability index for DDFL}
    \small
    \begin{tabular}{c|ccc|ccc|ccc}
        \hline
        \textbf{} & \multicolumn{3}{c}{ \textbf{Approach 01 $\gamma$ = 0.1  $\lambda$ = 0.8}} \vline & \multicolumn{3}{c}{ \textbf{Approach 02 $\gamma$ = 0.1 $\lambda$ = 0.9}} & \multicolumn{3}{c}{ \textbf{Approach 03 $\gamma$ = 0.2 $\lambda$ = 0.9}}\\
        \textbf{b}& \textbf{MNIST} & \textbf{CIFAR-10} & \textbf{CIFAR-100} & \textbf{MNIST} & \textbf{CIFAR-10} & \textbf{CIFAR-100} & \textbf{MNIST} & \textbf{CIFAR-10} & \textbf{CIFAR-100}\\ \hline
        10 & 97.62 & 89.22 & 83.38 & 97.76 & 88.74 & 81.83 & 97.60 & 88.62 & 82.06 \\
        25 & 97.21 & 88.40 & 82.71 & 97.71 & 87.97 & 82.27 & 97.19 & 87.68 & 81.69 \\
        50 & 96.85 & 87.85 & 82.10 & 97.05 & 87.55 & 81.63 & 96.57 & 87.06 & 80.89 \\
        100 & 96.24 & 86.41 & 81.04 & 96.99 & 87.18 & 80.67 & 96.04 & 86.17 & 80.23 \\
        200 & 92.26 & 83.83 & 78.60 & 92.49 & 83.88 & 78.26 & 93.47 & 85.08 & 79.11 \\ \hline
    \end{tabular}
\end{table*}

\begin{table}
    \centering
    \caption{The reduction in the test accuracy \textit{FedAvg} for non-IID data against MINST, CIFAR-10 and CIFAR-100 datasets}
    \centering
    \begin{tabular}{|c|c|c|c|}
    \hline
     \textbf{Data Distribution} & \textbf{MNIST} & \textbf{CIFAR-10} & \textbf{CIFAR-100}\\
    \hline
    IID FedAvg (Baseline) & 0.98 & 4.18 & 8.25 \\
    \hline
    non-IID FedAvg & 11.67 & 30.91 & 40.87 \\
    \hline
    \end{tabular}
    \label{Table: Accuracy Loss for MNIST, CIFAR-10 and CIFAR-100}
\end{table}

\subsection{Model reliability index: }
The reliability index is a crucial metric for evaluating the performance of ML systems. In this study, we calculate the reliability index, denoted as $\zeta_k$ and defined in \cite{AccurateDib2018}, as the ratio of the standard deviation of the test accuracy ($\sigma_n$) to the mean value of the test accuracy ($\mu_n$) per batch., expressed as Equation (\ref{eq:reliability_index}): Moreover, we have used MNIST, CIFAR-10 and CIFAR-100 dataset calculate the reliability index.

\begin{equation}
    \zeta_k (\%) = \left(1 - \frac{\sigma_n}{\mu_n}\right) \times 100
    \label{eq:reliability_index}
\end{equation}
    
Subsequently, the overall system reliability index ($\zeta_{\text{system}}$) can be computed by averaging all the reliability indexes, as shown in Equation (\ref{eq:system_reliability_index}):

\begin{equation}
    \zeta_{\text{system}} = \frac{1}{N} \sum_{k=1}^{N} \zeta_k
    \label{eq:system_reliability_index}
\end{equation}
    
\textbf{Table. V} quantifies the computed reliability index per batch size and illustrates the overall system stability. It is observed that Precision-weighted FL achieves optimal performance in most cases, except for CIFAR-100. This sensitivity to small batch sizes compromises the performance, as highlighted in the table. Moreover, we have found that the overall system reliability index is \textbf{88.11\%}

We state that the proposed approach reduces the bias term in FL with non-IID data. Let \(\delta_k(n)\) denote the bias term for device \(k\) at epoch \(n\), representing the difference between the local model ${F_k(\mathbf{w})}$and the globally optimal model ${F_k^n(\mathbf{w})}$.

In Dynamic Data distribution the  effect of dynamic data distribution on the bias term can be denied by,
\begin{equation}
    \delta_k(n+1) = \delta_k(n) + \epsilon_{\text{{distribution}}}
\end{equation}

This will lead to the normalized class distribution of each device and affect the bias term such that, 
\begin{equation}
    \delta_k(n+1) = \delta_k(n) + \epsilon_{\text{{normalization}}}
\end{equation}

Furthermore, the effect of entropy-based device selection on the bias term can be denoted as,
\begin{equation}
    \delta_k(n+1) = \delta_k(n) + \epsilon_{\text{{entropy}}}
\end{equation}

Finally, the  overall impact can be denoted as Combining the effects of the bias term:
\begin{equation}
    \delta_k(n+1) = \delta_k(n) + \epsilon_{\text{{distribution}}} + \epsilon_{\text{{normalization}}} + \epsilon_{\text{{entropy}}}
\end{equation}

The results indicate that the proposed approach converges well against the non-IID data setting These effects collectively contribute to reducing the bias term (\(\delta_k(n+1) < \delta_k(n)\)), indicating that the proposed approach leads to a reduction in bias over epochs. Moreover, we can state that the accuracy increment and faster convergence signify the global model moves away from the sub-optimal models.

%%%%%%%%%%%%%%%%%%%%%%%%%%%%%%%%%%%%%%%%%%
%%%%%%%%%%%%%%%%%%%%%%%%%%%%%%%%%%%%%%%%%%

\section{Conclusion}
\label{Section: Conclusion}

This research has delved into the intricacies of statistical challenges in FL, particularly prevalent in scenarios with non-IID data. Our observations unveiled a substantial accuracy reduction, rooted in weight divergence and characterized by the bias term (\(\delta_k\)). To counteract these challenges, our innovative dynamic data distribution mechanism dynamically allocates a global subset of data, fostering faster convergence and a remarkable enhancement in accuracy. Reducing the bias term signifies a pivotal stride towards more robust model convergence.
Additionally, this dynamic distribution significantly reduces data skewness, ensuring a more representative and balanced training process. Our approach also addresses the fairness of device selection during model aggregation through an entropy-based aggregation mechanism. By leveraging data entropy metrics, we ensure that the device selection process is unbiased and representative, contributing to the overall fairness and reliability of the FL process. Furthermore, we also provided a theoretical upper bound to guarantee convergence. Quantitatively, our experiments demonstrated a substantial accuracy boost of approximately 5\% for the MNIST dataset, around 18\% for CIFAR-10, and around 20\% for CIFAR-100, all achieved with only a 10\% global subset of data. This accuracy increment, coupled with faster convergence, and low computational overhead highlights the practical efficacy of our proposed methodologies.

\bibliographystyle{IEEEtran}
\bibliography{references}
\vspace{-0.5cm}
\begin{IEEEbiography}[{\includegraphics[width=1in,height=1.25in,clip,keepaspectratio]{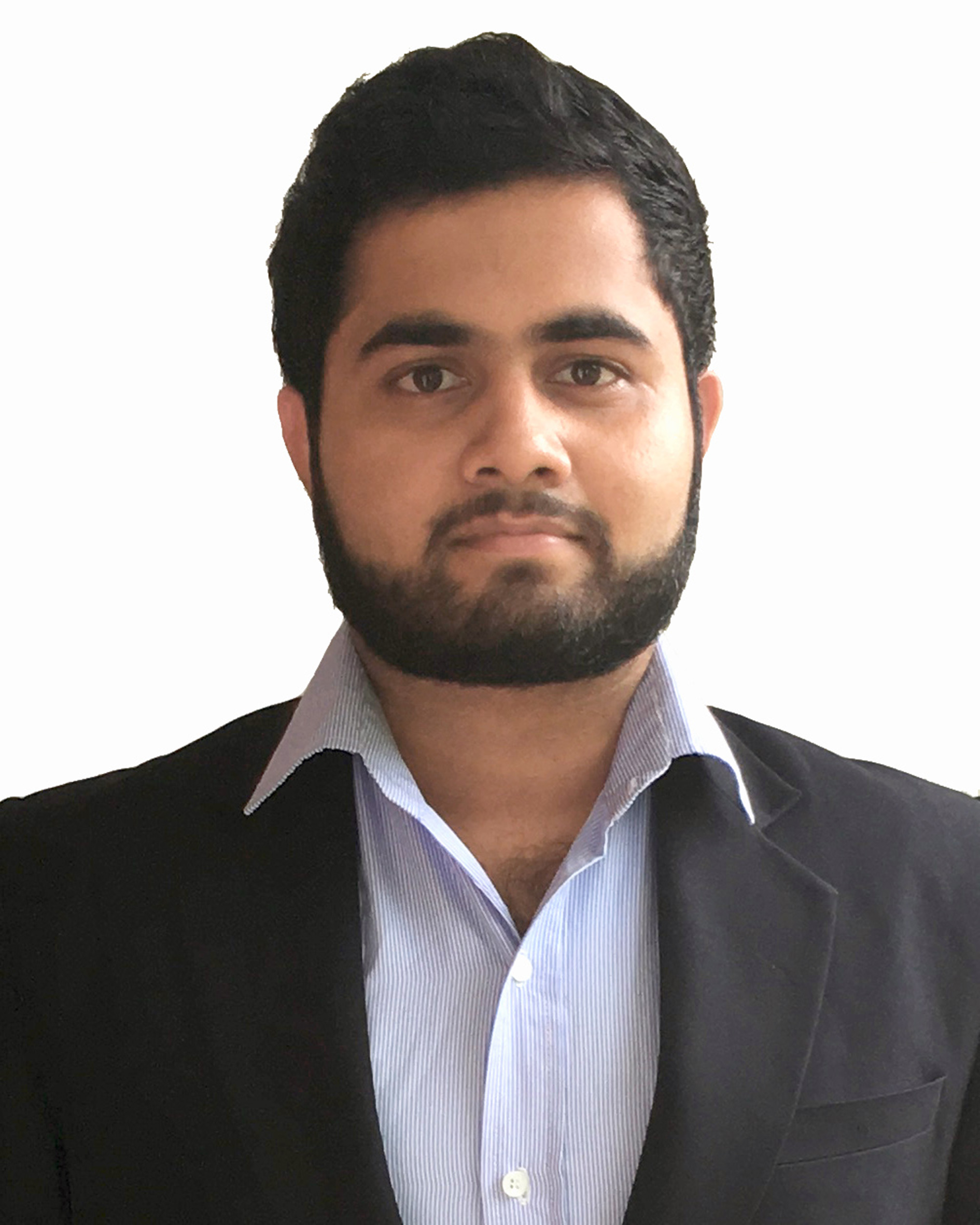}}]{Charuka Herath}  reading the PhD degree in AI and Cyber security at Loughborough University, London and an alumnus of the University of Moratuwa, Sri Lanka. Previously, he worked as a Senior Software Engineer for Sysco LABS, and WSO2 Private Limited respectively, from 2020 to 2022. He is an Associate lecturer in computing at Arden University, London. His research interests include Wireless distributed learning, one-shot learning, and Cyber security. 
\end{IEEEbiography}
\begin{IEEEbiography}[{\includegraphics[width=1in,height=1.25in,clip,keepaspectratio]{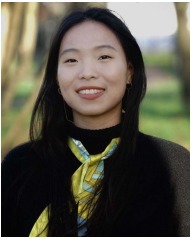}}]{Xiaolan Liu}   received the Ph.D. degree from the Queen Mary University of London, U.K., in 2021. She was a Research Associate with King’s College London from 2020 to 2021. Since October 2021, she has been with the Institute
for Digital Technologies, Loughborough University London, U.K., where she is a Lecturer. Her
current research interests include wireless distributed learning, multi-agent reinforcement learning for
edge computing, and machine learning for wireless communication optimization.
\end{IEEEbiography}
\begin{IEEEbiography}[{\includegraphics[width=1in,height=1.25in,clip,keepaspectratio]{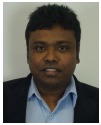}}]{Sangarapillai Lambotharan} received the PhD degree in signal processing from Imperial College London, U.K., in 1997. He was a Visiting Scientist with the Engineering and Theory Centre, Cornell University, USA, in 1996. Until 1999, he was a Post-Doctoral Research Associate at Imperial College London. From 1999 to 2002, he was with the Motorola Applied Research Group, U.K., where he investigated various projects, including physical link layer modelling and performance characterization of GPRS, EGPRS, and UTRAN. He was with King’s College London and Cardiff University as a Lecturer and a Senior Lecturer, respectively, from 2002 to 2007. He is a Professor of Signal Processing and Communications and the Director of the Institute for Digital Technologies, at Loughborough University London, U.K. His current research interests include 5G networks, MIMO, blockchain, machine learning, and network security. He has authored more than 250 journal articles and conference papers in these areas. He is a Fellow of IET and a Senior Member of IEEE. He serves as an Associate Editor for the IEEE Transactions on Signal Processing and IEEE Transactions on Communications.  
\end{IEEEbiography}
\begin{IEEEbiography}[{\includegraphics[width=1in,height=1.25in,clip,keepaspectratio]{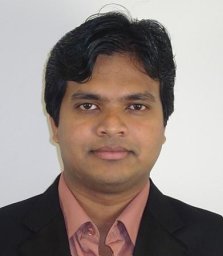}}]{Yogachandran Rahulamathavan} is a Reader in Cyber Security and Privacy and joined Loughborough University London in 2016. After obtaining a Ph.D. in Signal Processing from Loughborough University in 2012, Rahul joined the Information Security Group at City, University London as a Research Fellow to lead signal processing in the encrypted domain research theme. During his time as a research fellow, he was a security and privacy work package leader for a Large-Scale Integrated Project SpeechXrays funded by the European Commission. Since April 2016, he joined Loughborough University's postgraduate campus in London as a lecturer, where he was promoted to senior lecturer in January 2020 and to Reader in January 2024. He is the module leader for Cybersecurity and Forensics, Principles of Artificial Intelligence and Data Analytics and Information Management modules. He is one of the recipients of the British Council's UK-India research funding in 2017 and successfully led a project between Lboro, City and IIT Kharagpur. Currently, Rahul leads a team of three Ph.D. students and serves as the principal investigator for an industry-funded project supported by Airbus Defense. Rahul is a Programme Director for MSc Cyber Security and Data Analytics at Loughborough University London.
\end{IEEEbiography}

\end{document}